\documentclass[journal]{IEEEtran}
\usepackage{cite}
\usepackage{multicol}
\usepackage{multirow}
\usepackage{amsmath}
\usepackage{amssymb}
\usepackage{balance}
\usepackage{url}
\usepackage{nameref}
\usepackage{hyperref}
\usepackage[table]{xcolor}
\usepackage[pdftex]{graphicx}
\usepackage{ifpdf}
\ifpdf
  \DeclareGraphicsExtensions{.pdf,.png,.jpg}
\else
  \DeclareGraphicsExtensions{.eps}
\fi

\usepackage{array}
\usepackage[caption=false,font=footnotesize]{subfig}

\makeatletter
\newcommand{\mypm}{\mathbin{\mathpalette\@mypm\relax}}
\newcommand{\@mypm}[2]{\ooalign{%
        \raisebox{.25\height}{$#1+$}\cr
        \smash{\raisebox{-.5\height}{$#1-$}}\cr}}
\makeatother

\newtheorem{defn}{Definition}
\newcommand{\R}{{\mathbb R}} 
\newcommand{\N}{{\mathbb N}} 
\newcommand{\E}{{\mathbb E}} 
\newcommand{\No}{{\mathcal N}}

\newcommand{\x}{\,\mathbf x}

\usepackage{pgfplots}
\pgfplotsset{compat=1.13} % added on LATEX suggestion
\definecolor{lg}{rgb}{0.8500,0.3250,0.0980}
\definecolor{lr}{rgb}{0.9290,0.6940,0.1250}

\begin{document}
\bstctlcite{AuthorControl}
\title{A Multi-layer Gaussian Process for Motor Symptom Estimation in People with Parkinson's Disease}

\author{%Michael~Shell,~\IEEEmembership{Member,~IEEE,}
        %John~Doe,~\IEEEmembership{Fellow,~OSA,}
        %and~Jane~Doe,~\IEEEmembership{Life~Fellow,~IEEE}% <-this % stops a space
        Muriel~Lang\textsuperscript{1*},~\IEEEmembership{Member,~IEEE,}
		Franz~M.~J.~Pfister\textsuperscript{2},	
        Jakob~Fr\"ohner\textsuperscript{1},
        Kian~Abedinpour\textsuperscript{3,4},
        Daniel~Pichler\textsuperscript{3,4},
		Urban~Fietzek\textsuperscript{3},
        Terry~Taewoong~Um\textsuperscript{5},
		Dana~Kuli\'{c}\textsuperscript{5},~\IEEEmembership{Member,~IEEE,}
		Satoshi~~Endo\textsuperscript{1},
		and~Sandra~Hirche\textsuperscript{1*},~\IEEEmembership{Senior Member,~IEEE,}
\thanks{\textsuperscript{1} Department of Electrical and Computer Engineering, Technical University of Munich, Munich, Germany.}%
\thanks{\textsuperscript{2} Faculty of Mathematics, Informatics and Statistics, Ludwig-Maximilians-Universit\"{a}t M\"{u}nchen, Munich, Germany.}%
\thanks{\textsuperscript{3} Department of Neurology and Clinical Neurophysiology, Sch\"{o}n Klinik M\"{u}nchen Schwabing, Munich, Germany.}%
\thanks{\textsuperscript{4} Department of Neurology, School of Medicine, Technical University of Munich, Munich, Germany}%
\thanks{\textsuperscript{5} Department of Electrical and Computer Engineering, University of Waterloo, Waterloo, ON, Canada.}%
\thanks{\textsuperscript{*} email: \{muriel.lang, hirche\}@tum.de}% <-this % stops a space
\thanks{Manuscript received \textit{date}; revised \textit{date}.}} %April 19, 2005

%\markboth{Transactions on Biomedical Engineering,~Vol.~?, No.~?, date?}%
%{Lang \MakeLowercase{\textit{et al.}}: Ambient symptom recognition of Parkinson's disease from wearable tracking device}

\maketitle

\begin{abstract}
The assessment of Parkinson's disease (PD) poses a significant challenge as it is influenced by various factors which lead to a complex and fluctuating symptom manifestation. Thus, a frequent and objective PD assessment is highly valuable for effective health management of people with Parkinson's disease (PwP). Here, we propose a method for monitoring PwP by stochastically modeling the relationships between their wrist movements during unscripted daily activities and corresponding annotations about clinical displays of movement abnormalities. We approach the estimation of PD motor signs by independently modeling and hierarchically stacking Gaussian process models for three classes of commonly observed movement abnormalities in PwP including tremor, (non-tremulous) bradykinesia, and (non-tremulous) dyskinesia. We use clinically adopted severity measures as annotations for training the models, thus allowing our multi-layer Gaussian process prediction models to estimate not only their presence but also their severities. The experimental validation of our approach demonstrates strong agreement of the model predictions with these PD annotations. Our results show the proposed method produces promising results in objective monitoring of movement abnormalities of PD in the presence of arbitrary and unknown voluntary motions, and makes an important step towards continuous monitoring of PD in the home environment.
\end{abstract}

\begin{IEEEkeywords}
Ambient intelligence, Gaussian processes, Medical information system, Wearable sensors
%Enter key words or phrases in alphabetical order, separated by commas. For a list of suggested keywords, visit \url{http://www.ieee.org/organizations/pubs/ani_prod/keywrd98.txt}
\end{IEEEkeywords}

\IEEEpeerreviewmaketitle
% alter predefined IEEE bib style format by calling "nocite" --> options are first entry in bibfile "bib"
\bstctlcite{IEEEexample:BSTcontrol}

%%%%%%%%%%%%%%%%%%%%%%%%%%%%%%%%%%%%%%%%%%%%%%%%%%%%%%%%%%%

\section{Introduction}
\IEEEPARstart{P}{arkinson's} disease (PD) is the second most common neurodegenerative disease after Alzheimer's disease, largely associated with various forms of movement-related symptoms~\cite{Robertson1990}. As many as~60,000 new cases are diagnosed every year in North America~\cite{DeLau2006}, and according to a recent analysis from~\cite{Enders2017}, the prevalence of PD is estimated to be 217.22/100,000 in Germany. 
Typically, people with PD (PwP) show several characteristic movement deficits, such as brady-hypokinesia, rigidity, tremor, postural instability, and movement initiation disorder (``freezing'')~\cite{Hoehn1967,Fahn1995,Hughes1992}. 
Brady-hypokinesia is the most salient symptom and is characterized by slowness and reduction of movements~\cite{Berardelli2001}. Rigidity is a stiffening of the body parts~\cite{Hallett2003}. Tremor typically occurs as a rest tremor, and is defined as an involuntary rhythmical muscle contraction of 4-6~Hz frequency~\cite{DeuschlGuntherFietzek2003}, which is a common initial PD symptom~\cite{Hoehn1967}. 
%The Movement Disorder Society sponsored Unified PD Rating Scale (MDS-UPDRS), has been developed to assess the these characteristic motor functions of people with PD~\cite{Goetz2007,Goetz2008}. The MDS-UPDRS is a 5-level rating scale and assesses motor symptoms, among others, and it relies on patient-reported data and assessments by experts on movement disorders. By large, clinical trials on neurorehabilative measures use the motor part of the MDS-UPDRS to establish clinical efficacy of any given intervention~\cite{angeles_automated_2017,butt_leap_2017,Mirelman2011,Papapetropoulos2010,Summa2017,Esculier2012}.
As the aetiology of PD lies in dopamine deficiency due to the degeneration of substantia nigra, these PD symptoms are controlled primarily by dopaminergic treatments~\cite{Surmeier2017}. However, as a side effect, many patients may exhibit dyskinesia, characterized by involuntary muscle movements~\cite{Jankovic2005}. \\
PD exhibits symptom heterogeneity between patients, especially in the advanced stages of the disease~\cite{McColl2002}. In addition, the short-term influence of medication, and the dependency on a variety of internal and external factors leads to highly complex and variable displays of the PD symptoms. For these reasons, even for highly trained clinicians, there is considerable inter-rater and intra-rater variability in judging the severity of the cardinal symptoms~(e.g.,~\cite{Heldman2011,Post2005}), which compromises both diagnosis and monitoring of PD. 
Thus, frequent and objective clinical assessments are important for deriving a suitable health management plan (e.g., adapting drug dosage).
Nevertheless, PD assessments in a patient's home environment are often impractical due to time and economic constraints. An alternative is self-disclosure by the patients in a diary which is reported on a regular basis throughout the day. However, the reliability of those diaries is often questionable as these reports are subjective and it requires substantial commitment by the patients~\cite{Antonini2011}. %and possible false statements can occur from variation in the patients' training and interpretation of symptom manifestation, confusion of symptoms and indication locations or the inability of patients to fill in the diary on a regular basis. 
Therefore, we propose a supervised machine learning method for autonomously estimating and monitoring PD-specific movement abnormalities using a commercial wrist-worn wearable sensor to track the patients' motion.
These wearable devices can be used to continuously monitor the movement of the users with minimum intrusion on their daily activities. In free living settings, however, tracked movements are often comprised of motor expressions of PD as well as unknown movements related to daily activities. Thus, the learning algorithm must be able to attend to the features of PD symptoms in the face of highly variable inputs.\\
In the present study, therefore, we estimate the severity of PD-specific movement abnormalities in the presence of arbitrary and unknown voluntary motions by learning the movement features including those in the time-frequency domain. Specifically, we use a multi-layered Gaussian process (GP) model, in which each of the hierarchically stacked GP models estimates severity of different PD-related movement abnormality classes, namely \emph{tremor} (bradykinetic or dyskinetic), non-tremulous \emph{bradykinesia} and non-tremulous \emph{dyskineasia}.

\subsection{Related Work}
%Numerous supervised machine learning techniques are available in literature, and we provide an overview of the relevant subset that addresses autonomous Parkinson's symptom detection and recognition in the following. 
Machine learning has been employed in many studies to model the relationships between movements of PwP and severity of PD~\cite{angeles_automated_2017, eskofier_recent_2016, ghassemi_combined_2016, cancela_comprehensive_2010, roy_resolving_2011}. 
Commonly, the movements are recorded using wearable sensors such as accelerometer, gyroscope, electromyography and video-based tracking devices. In the majority of studies (e.g.~\cite{eskofier_recent_2016, angeles_automated_2017, cancela_comprehensive_2010, zwartjes_ambulatory_2010}), wearable inertial measurement units (IMUs) are used with applications in remote home environments in mind, whereas marker-based motion tracking systems are often used in a controlled environment or when they provide a ground truth for other sensors~\cite{kubota_machine_2016, filippeschi_survey_2017}.
Existing studies largely focus on one or two PD-specific behavior classes including tremor, bradykinesia, dyskinesia, and gait disturbance, where the presence of abnormality in these classes is estimated by machine learning~\cite{kubota_machine_2016}. %We, however, aim for a comprehensive solution for PD motor symptom detection.
As labeled training data is necessary for supervised machine learning, a movement disorder specialist commonly monitors the patients and labels clinical observations according to the Movement Disorder Society sponsored Unified PD Rating Scale (MDS-UPDRS) with which motor symptoms are assessed on a 5-level rating scale~\cite{Goetz2007,Goetz2008}. 

% To ensure unsupervised usability by patients, only wearable inertial modules (accerlerometer and gyroscope) can be taken into consideration, as only those allow for prompt usage without complex calibration or adjustment routine.

For instance, Eskofier et al.~\cite{eskofier_recent_2016} compare several supervised machine learning pipelines and a deep learning algorithm to detect bradykinesia. Several specific motor tasks of 10 patients with idiopathic PD were recorded using IMU sensors. Every task was rated by a movement disorder specialist according to the MDS-UPDRS rating scale. Using these data, a classification accuracy of up to 85\% was achieved with standard machine learning techniques such as support vector machine (SVM) while deep learning demonstrated 90\% accuracy in predicting the presence and absence of bradykinesia. 
Angeles et al.~\cite{angeles_automated_2017} were also able to classify PD symptoms such as kinetic tremor according to the MDS-UPDRS score given by clinicians with an accuracy of up to 87\% using simple tree, linear SVM and fine k-nearest neighbor (kNN) algorithms. An accuracy of up to 92\% was achieved in predicting bradykinesia using Fine kNN. 
As an alternative to wearable sensors, video-based methods have been used to investigate the PD motor symptoms. 
Butt et al.~\cite{butt_leap_2017, wahid_classification_2015}, for example, used the hand motion data collected using an RGB-D camera to distinguish PD patients from healthy individuals at an accuracy of up to 85\% using SVM. 
Alternatively, a semi-supervised classification algorithm based on k-means and self-organizing tree map clustering was applied in~\cite{Hssayeni2016}, obtaining accuracies in the range of 42-99\% for patients with different levels of dyskinesia severity. In a hybrid approach,~\cite{tzallas_perform:_2014} tailors a different set of classifiers including the hidden Markov model, decision tree, SVM, and random forest to estimate the tremor, dyskinesia, bradykinesia and freezing of gait, respectively. Using the IMU signals collected from four locations on the body and the patient's own diary rating as class labels, the authors predicted the severity of PD symptoms with above 70\% accuracy during short-term scripted activities.

The experiments for modeling motor aspects of PD described so far took place in laboratory environments, where the patients performed scripted activities. Few examples are available on symptom modeling during free living in the literature.
In~\cite{Hoff2004} binary classification of bradykinesia and tremor attained 60-71\% sensitivity using a receiver operating characteristic curve on single kinematic variables. 
In contrast, using a neural network with 23 kinematic variables from 6 IMU sensors, Keijsers et al.~\cite{Keijsers2006} achieved an average sensitivity and specificity of 97\% for binary classification. %In contrast, a single waist sensor was used in a long-term (1-3 days) monitoring experiment by~\cite{Perez-Lopez2016} in which specificity and sensitivity of binary classification were higher than 90\%.
Supervised machine learning is used in~\cite{cancela_comprehensive_2010} to classify bradykinesia during free movement and daily living. The authors contrasted a set of classifiers such as SVN and k-means and reported 70\% to 86\% classification accuracy for predicting the bradykinesia severity. 
%
%The output results were compared against the patient's own diary rating and a classification accuracy of 87\% was achieved in predicting the presence and absence of tremor using hidden Markov model. Predicting the presence and absence of bradykinesia, a classification accuracy of approximately 75\% was achieved using SVM. Hence, again in this approach, symptom classification was binary.
%Furthermore, the technology is restricted by specific hardware requirements. For the presented results, the PERFORM systems required four triaxial accelerometer devices to be physically attached the PD patient's body at each extremity, one accelerometer/gyroscope sensor on the waist and one data acquisition unit.

Existing studies largely focus on modeling PD symptoms either in a controlled environment and/or using a binary classification algorithms due to the limited PD data set and a highly variable display of PD symptoms. Thus, the estimation of the symptom severity has been challenging task, and previous works have not sufficiently explored the benefits of advanced machine learning techniques such as Gaussian mixture modeling and Gaussian process regression with which non-linear functions can be robustly approximated with relatively small datasets.

\subsection{Contributions and Article Structure}
Our contributions in improving severity estimation of behavioral abnormalities of PwP during unscripted daily activities are supported by two key methodological achievements: (1) identification of appropriate data features for our PD classes~(\ref{sec:feature}) and (2) the hierarchical structured machine learning models which handle each PD class~(\ref{sec:MGP}). We propose an approach to autonomously estimate severities in three classes of PD-related motor abnormalities (i.e., PD classes); namely tremor, and non-tremulous episodes of bradykinesia and dyskinesia. As the tremor motion is highly characteristic and easily differentiable from other motions, the model estimates the tremor presence and its severity in the first layer, based on the frequency ratio of the tremor motion versus voluntary actions. Subsequently, in the second layer, the remaining data are used to analyze the severity of bradykinesia and dyskinesia. 
%The absence of movement disorders (i.e., balanced state) is assumed when no sign of these three PD classes is present. 
For each of the three class estimates, we apply GP regression and prediction and thus obtain a multi-layer GP model. Our approach focuses on collections of unscripted motion data, thus the PD estimation can take place during unconstrained free living activities. We purposefully use a single commercially available wrist-worn low-cost wearable sensor to demonstrate robustness and applicability of our approach to real-life scenarios.

The remainder of the paper is structured as follows. We first introduce our methods for data acquisition. Second, we present our modeling approach, consisting of feature identification, hierarchical symptom estimation and the application setup. Then, we present our experimental results. Finally, we discuss the results and conclude the paper.

%%%%%%%%%%%%%%%%%%%%%%%%%%%%%%%%%%%%%%%%%%%%%%%%%%%%%%%%%%%%%%%%%%%%%%%%%%%%%%%%
\section{Data Collection}
In this section we describe how the patient data was collected, introducing the patient cohort of our study, the sensor device and the setting for data capture. Subsequently, the procedure of data acquisition is explained.%, which was performed during unrestricted daily-living activities.

Thirty individuals who took part in the study were diagnosed with PD by a neurologist according to the UK Brain Bank Diagnostic Criteria~\cite{Gibb1989} at the Sch\"on Klinik M\"unchen Schwabing, Germany. The average age of the participants was 67$\mypm$10, and 20 were male and 10 were female. The mean disease duration was 11$\mypm$5 years. 
The median of the patients' disease progress according to the Hoehn and Yahr scale~\cite{Hoehn1967} is 3.5 with an interquartile range of 1.
%All patients were chronically treated with Levadopa preparations and their mean levodopa equivalent dose on the day of testing was {\color{red}[NUMBER $\pm$ NUMBER]}, calculated according to~\cite{Tomlinson2010}. 
The recruitment of the patient cohort and the data acquisition were performed at the Sch\"on Klinik M\"unchen Schwabing (Munich, Germany). This study was approved by the ethical board of the Technical University of Munich~(Ref. No. 234/16S). 

In order to learn and predict the display of movement abnormalities using GP, the movements of the participants were recorded, together with corresponding PD annotations including their class and severity.
The linear acceleration and angular velocity of the wrist were measured using the Microsoft Band~2 (Microsoft). Inside the band is a 6-axis gyroscope/accelerometer module~(LSM6D series by STMicroelectronics) and a bluetooth communication module~(Bluetooth~4.0) for transmitting the data to a peripheral device. The accelerometer registers motion up to $\mypm 8~G$~($G = 9.81~m/s^{2}$) with sensitivity of 0.244~$mG/LSB$~(least significant byte). The data range of the gyroscope is $\mypm$1000~$dps$ (degree per second) with sensitivity of 35~$mdps/LSB$. An Android application, "MS band data collector (pro)" was used to stream and store the data on a Samsung Galaxy A5~(Android 6.0.1). 

Observations of the PD classes and their severities were concurrently performed every minute during the data collection period by a trained expert (D.P.) who accompanied PwP and passively monitored them.
The symptom labels for bradykinesia and tremor were performed according to the MDS-UPDRS in a standard 5-level rating scale, where they correspond to item III.14 and III.17, respectively. Dyskinesia was assessed using the abnormal involuntary movement scale (AIMS, item A2.5). For all these three types of motor signs, the absence of abnormality is rated as 0 and the severity levels correspond to 1 = slight, 2 = mild, 3 = moderate and 4 = severe. When none of the abnormalities were present, the motor state of the patients was considered to be eukinetic, and classed as \emph{balanced}. Furthermore, voluntary activities (e.g., walking, standing, lying/resting and sitting) were reported in the same one-minute time window. When multiple types of the motor signs and activities were present within the same time window, the predominant motor sign and activity were reported.
The data collection was performed in a free-living environment during the regular in-patient stay at the hospital for medication adaptation. On average, the data were collected for 331.2$\mypm$192.6 minutes per participant, totaling 9937.0 minutes across the participants. After being briefed on the procedure, the participants wore the band on the wrist of the more affected side. Once the bluetooth connection was established between the band and smartphone which locally stored the data, the participants were free to engage in any daily routine including activities outside of the hospital. The recording ended when the patients desired, or before going to bed at the latest. Furthermore, the sensing device was disabled when the patients were in the toilet/bath or when requested.

%%%%%%%%%%%%%%%%%%%%%%
\section{Hierarchical PD Symptom Recognition}
In this section, the approach for autonomous PD detection and estimation are presented. 
We start by characterizing the collected data and their relevance for estimating PD classes, thereby motivating a selection of motion features. 
The PD classes are then assessed by a multi-layer GP model.
In the first layer, a GP estimates the presence of tremor and its severity. In case tremor is absent, the second layer is triggered, where two estimations are performed in parallel; one for dyskinesia and the other for bradykinesia. Hence, the severity level of both movement symptoms is predicted for all incoming data in the second layer. 

Then, we select the symptom class with the larger predicted value, as we intend to estimate the predominant motor symptom.
The absence of movement disorders (i.e., balanced state) is assumed when no sign of these three PD classes is present.

\subsection{Data Analysis}
\label{data}
\begin{figure}[!t]
    \centering
    \includegraphics[width=\columnwidth]{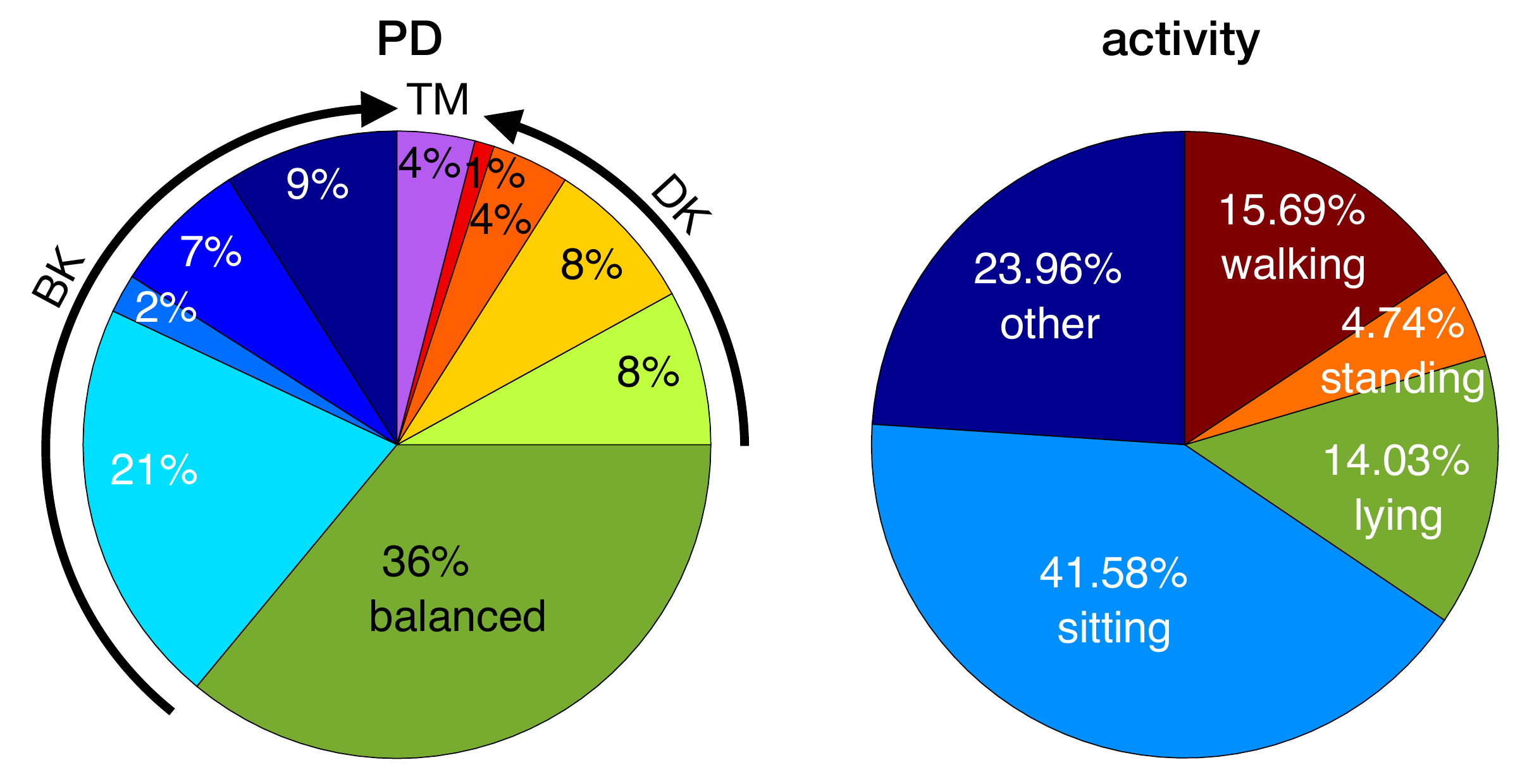}
    \caption{{\bf Distribution of PD class severities and activities of the participants.} The label distributions were calculated from the collected dataset for each participant and then averaged. The arrows in the PD class chart (left) indicate increasing severity of dyskinesia (DK) and bradykinesia (BK) scores in 4 levels each. The tremor class (TM) also has four severity levels, but is visualized here as a single section (purple).}
    \label{symp_fig}
\end{figure}

%In the following we provide an overview over the collected patient data characteristics.
%The descriptive analysis of the symptom labels, collected by the clinical expert, shows that $46.3\%$ belongs to the balanced class, and the rest are distributed equally between the bradykinesia (BK) class with $29.4\%$ and dyskinesia (DYS) class with $24.3\%$, see Fig.~\ref{pie}. The labeled data also suggests that patients spent large proportion of time sitting on a chair ($44.4\%$). The activities gathered in category \textit{other} comprise specific tasks, as for instance eating, climbing stairs and brushing teeth or activities with external acceleration like taking a train or elevator. 
In the following, we provide an overview of the collected patient data.
The descriptive analysis of the PD annotations shows that 35.95\% belongs to the balanced class, while the bradykinesia and dyskinesia were observed in 38.70\% and 21.13\% of the data, respectively. The percentages are normalized between participants as the data size differs between them. The annotations also indicate patients spent a large proportion of their time sitting on a chair (41.58\%), see Fig.~\ref{symp_fig} for details. The activities gathered in the category \textit{other} comprise specific tasks, for instance eating, climbing stairs and brushing teeth or activities that externally confound inertia measurements such as taking a train or being in an elevator. 

%In the following we provide an overview over the collected patient data characteristics.
%The descriptive analysis of the symptom labels, collected by the clinical expert, shows that $46.3\%$ belongs to the balanced class, and the rest are distributed equally between the bradykinesia (BK) class with $29.4\%$ and dyskinesia (DYS) class with $24.3\%$, see Fig.~\ref{pie}. The labeled data also suggests that patients spent large proportion of time sitting on a chair ($44.4\%$). The activities gathered in category \textit{other} comprise specific tasks, as for instance eating, climbing stairs and brushing teeth or activities with external acceleration like taking a train or elevator. 

%\begin{table}[!ht]
%    \begin{adjustwidth}{-0.8cm}{0in} %-1.5 Comment out/remove adjustwidth environment if table fits in text column.
%        \centering
%        \caption{
%            {\bf Distribution of symptom severities and activities of the participants.}}
%        \begin{tabular}{r+c c c c| c c c c|c c c c|c}
%        symptom & \multicolumn{4}{c|}{dyskinesia} & \multicolumn{4}{c|}{bradykinesia} & \multicolumn{4}{c|}{tremor} & balanced\\ 
%        severity & 1 & 2 & 3 & 4 & 1 & 2 & 3 & 4 & 1 & 2 & 3 & 4 & 0 \\ \hline
%        data $(\%)$ & 7.7 & 8.3  & 4.2 & 0.9 & 20.5 & 2.0  & 7.6 & 8.7  & 2.4 & 1.6 & 0.1 & 0 & 36.0\\
%         & \multicolumn{13}{c}{ }\\
%         activity & \multicolumn{2}{c}{sitting} & \multicolumn{2}{c}{laying} & \multicolumn{2}{c}{standing} & \multicolumn{2}{c}{walking} & \multicolumn{2}{c}{other} & \multicolumn{3}{c}{ } \\ \cline{1-11}
%         data $(\%)$ & \multicolumn{2}{c}{41.6} & \multicolumn{2}{c}{15.7} & \multicolumn{2}{c}{4.7} & \multicolumn{2}{c}{14.0} & \multicolumn{2}{c}{24.0} & \multicolumn{3}{c}{ }   \\
%    \end{tabular}
%        \begin{flushleft} The label distributions were calculated from the collected dataset for each participant and then averaged.
%        \end{flushleft}
%        \label{symp}
%    \end{adjustwidth}
%\end{table}

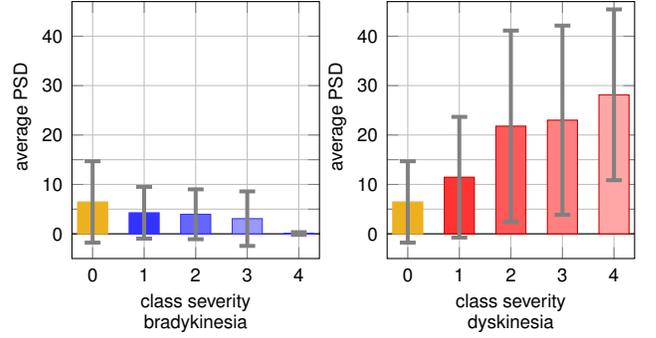
\begin{figure}[!t]
    \centering
    \subfloat{%
        \begin{tikzpicture}[font=\sffamily\scriptsize]
        %\definecolor{lb}{blue!40!white}
        \begin{axis}[%xscale = 1.45,
        %thick,
        width=0.55\columnwidth, height=5cm,
        %scaled ticks=false, 
        %anchor=north east,
        %tick label style={/pgf/number format/fixed},
        xlabel style={align=center},
        x label style={at={(axis description cs:0.5,-0.1)},anchor=north},
        %xbar, 
        xtick={1,2,3,4,5}, 
        xticklabels = {0,1,2,3,4}, 
        xlabel = {class severity\\ bradykinesia},
        grid=major,
        xmin = 1, xmax = 5,
        %ybar, 
        bar shift=0pt, 
        bar width=0.4cm, 
        enlarge x limits=0.1,
        ymin = -5, ymax = 47, 
        ylabel = average PSD,
        y label style={at={(axis description cs:-0.2,0.8)},anchor=east},
        ytick = {0,5,10,15,20,30,40}, 
        yticklabels={0,,10,,20,30,40},] 
        \draw[-, draw] (0,0) to (6,0);
        \addplot+[mark=none,color = blue!80, fill = blue!20, 
        ybar, solid,
        error bars/.cd, y dir=both, y explicit,
        error bar style={line width=1.5pt, gray, xshift=0cm},
        error mark options = {rotate = 90, line width=1.5pt, mark size = 3pt, gray,}]
        coordinates{(5,0.0631) +- (0,0.273595706904474)};
        \addplot+[mark=none,color = blue!80, fill = blue!40,%color = lr, fill = lr, 
        ybar, solid,error bars/.cd, y dir=both, y explicit,
        error bar style={line width=1.5pt, gray, xshift=0cm},
        error mark options = {rotate = 90, line width=1.5pt, mark size = 3pt, gray,}]
        coordinates{(4,3.073362068250275) +- (0,5.516763444889589)};
        \addplot+[mark=none,color = blue!80, fill = blue!60,%color = lg, fill = lg, 
        ybar, solid,error bars/.cd, y dir=both, y explicit,
        error bar style={line width=1.5pt, gray, xshift=0cm},
        error mark options = {rotate = 90, line width=1.5pt, mark size = 3pt, gray,}]
        coordinates{(3,3.958840302585484) +- (0,5.047889116290308)};
        \addplot+[mark=none,color = blue!80, fill = blue!80,%color = lg, fill = lg, 
        ybar, solid,error bars/.cd, y dir=both, y explicit,
        error bar style={line width=1.5pt, gray, xshift=0cm},
        error mark options = {rotate = 90, line width=1.5pt, mark size = 3pt, gray,}]
        coordinates{(2,4.266250140432615) +- (0,5.233076594048313)};
        \addplot+[mark=none,color = lr, fill = lr, %color = lg, fill = lg, 
        ybar, solid,error bars/.cd, y dir=both, y explicit,
        error bar style={line width=1.5pt, gray, xshift=0cm},
        error mark options = {rotate = 90, line width=1.5pt, mark size = 3pt, gray,}]
        coordinates{(1,6.461217535556324) +- (0,8.231480028356337)};
        \end{axis}
        \end{tikzpicture}}
    \subfloat{%
        \begin{tikzpicture}[font=\sffamily\scriptsize]%\footnotesize]
        %\definecolor{lb}{blue!40!white}
        \begin{axis}[%xscale = 1.45,
        %thick,
        width=0.55\columnwidth, height=5cm,
        %scaled ticks=false, 
        %anchor=north east,
        %tick label style={/pgf/number format/fixed},
        xlabel style={align=center},
        x label style={at={(axis description cs:0.5,-0.1)},anchor=north},
        %xbar, 
        xtick={1,2,3,4,5}, 
        xticklabels = {0,1,2,3,4}, 
        xlabel = {class severity \\ dyskinesia},
        grid=major,
        xmin = 1, xmax = 5,
        %ybar, 
        bar shift=0pt, 
        bar width=0.4cm, 
        enlarge x limits=0.1,
        ymin = -5, ymax = 47, 
        ylabel = average PSD,
        y label style={at={(axis description cs:-0.2,0.8)},anchor=east},
        ytick = {0,5,10,15,20,30,40}, 
        yticklabels={0,,10,,20,30,40},] 
        \draw[-, draw] (0,0) to (6,0);
        \addplot+[mark=none,color = lr, fill = lr, %color = lg, fill = lg, 
        ybar, solid,error bars/.cd, y dir=both, y explicit,
        error bar style={line width=1.5pt, gray, xshift=0cm},
        error mark options = {rotate = 90, line width=1.5pt, mark size = 3pt, gray,}]
        coordinates{(1,6.461217535556324) +- (0,8.231480028356337)};
        \addplot+[mark=none,color = black!20!red, fill = red!80, 
        ybar, solid,error bars/.cd, y dir=both, y explicit,
        error bar style={solid, line width=1.5pt, gray, xshift=0cm},
        error mark options = {rotate = 90, line width=1.5pt, mark size = 3pt, gray,}]
        coordinates{(2,11.463456247185030) +- (0,12.205822018768758)};
        \addplot+[mark=none,color = black!20!red, fill = red!65, 
        ybar, solid,error bars/.cd, y dir=both, y explicit,
        error bar style={solid, line width=1.5pt, gray, xshift=0cm},
        error mark options = {rotate = 90, line width=1.5pt, mark size = 3pt, gray,}]
        coordinates{(3,21.792863393697388) +- (0,19.346772525995650)};
        \addplot+[mark=none,color = black!20!red, fill = red!50, 
        ybar, solid, error bars/.cd, y dir=both, y explicit,
        error bar style={solid, line width=1.5pt, gray, xshift=0cm},
        error mark options = {rotate = 90, line width=1.5pt, mark size = 3pt, gray,}]
        coordinates{(4,23.004734240549325) +- (0,19.139123340159873)};
        \addplot+[mark=none,color = black!20!red, fill = red!35, 
        ybar, 
        solid,
        error bars/.cd, 
        y dir=both, 
        y explicit,
        error bar style={solid, line width=1.5pt, gray, xshift=0cm},
        error mark options = {rotate = 90, line width=1.5pt, mark size = 3pt, gray,}]
        coordinates{(5,28.137935307698860) +- (0,17.283268430286572)};
        \end{axis}
        \end{tikzpicture}}
    \caption{{\bf Average PSD of the non-tremulous accelerometer data per PD class severity.} The mean (visualized by bar height) of each PSD per class severity level with its standard deviation (gray error bar) is sorted from balanced to severe. In both figures the yellow bar depicts the average PSD level in the balanced state. On the left side the severity levels of bradykineasia (without tremor) are visualized in blue shades, on the right side the levels of dyskinesia in red shades. }
    \label{activity}
\end{figure}

As interaction of PD and activity annotations result in a wide bandwidth of motion intensities, we investigate the power spectral density (PSD) of the corresponding accelerometer data. The analysis shows that sitting and standing labels have PSD spread between 1-10~Hz and a clearly differentiated tremor at around 4-6~Hz. For the walking label, the characteristic tremor activity is partially absorbed by the walking frequency which has harmonics of $\approx$2.5~Hz, as walking itself generates a strong peak in PSD.
Lying is generally described by low power across the spectrum, meaning that arm motion occurs infrequently. The shift of the tremor PSD peak towards 7~Hz during lying might be caused by the change in the resonance frequency of the wrist due to a movement constrained by a lying surface~\cite{Lakie2012}.
\begin{figure}[!t]
    \centering
    \includegraphics[width=0.9\columnwidth]{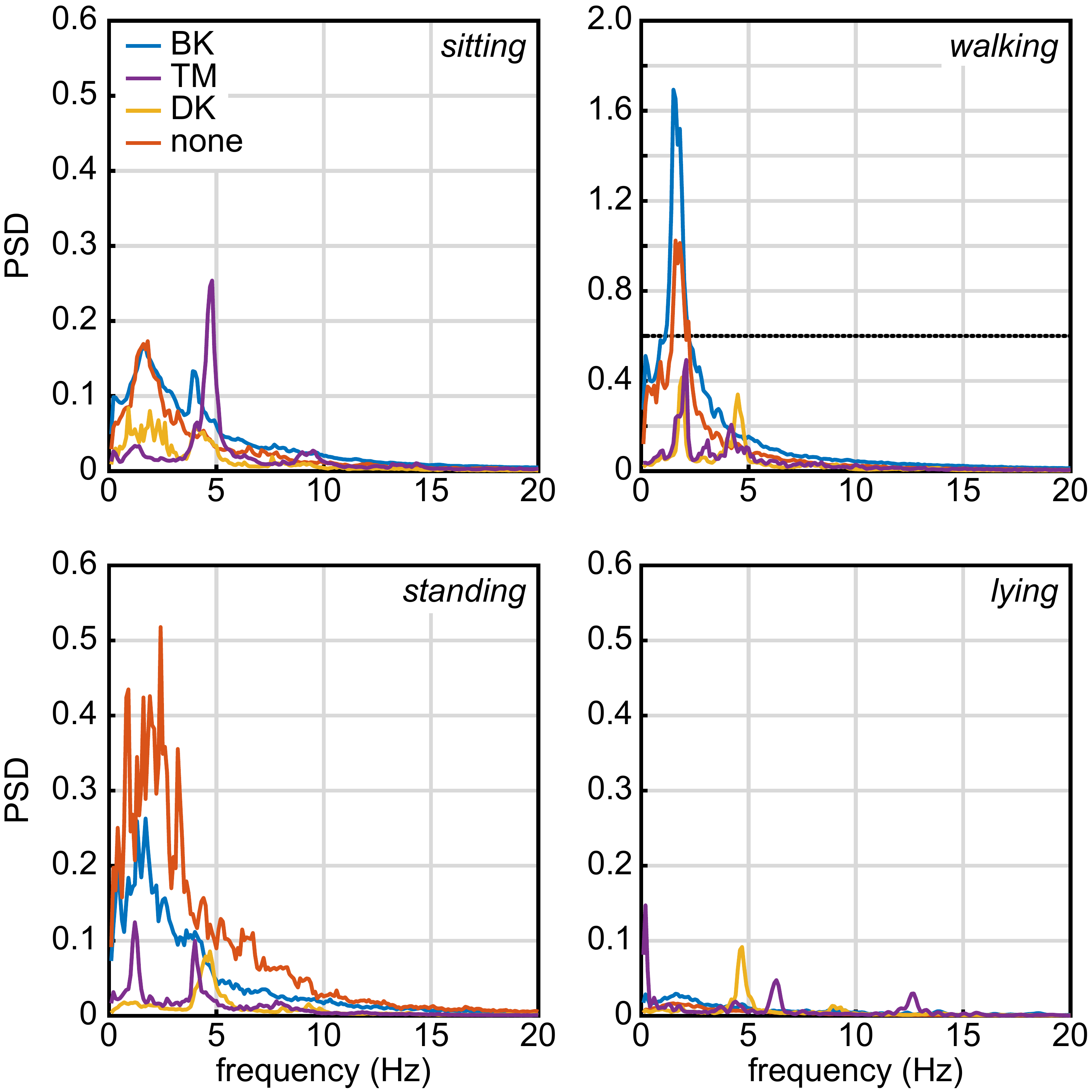}
    \caption{{\bf Power spectral density (PSD) of the accelerometer data.} The PSD is visualized for bradykinesia (BK), tremor (TM), dyskinesia (DK) and the balanced state (none). It shows generally high densities during walking, where the dotted line depicts the scale limit of the other three plots, and very low PSD values during lying.}
    \label{psd}
\end{figure}
Moreover, further analysis reveals on average increasing PSD 
%{\color{red}{[PDS is a power density across spectrum (Fig 2) - is this the power at the peak?}]} 
in the PD classes from bradykinesia through the balanced state to dyskinesia, as visualized in Fig.~\ref{activity}, even though the individual activities introduce a high noise level. The tremor class is omitted in this figure, as the PSD of tremor data tends to overshadow the other PD classes.

\subsection{Data Processing and Feature Generation}
\label{sec:feature}
\begin{figure}[!t]
    \centering
    \includegraphics[width=\columnwidth]{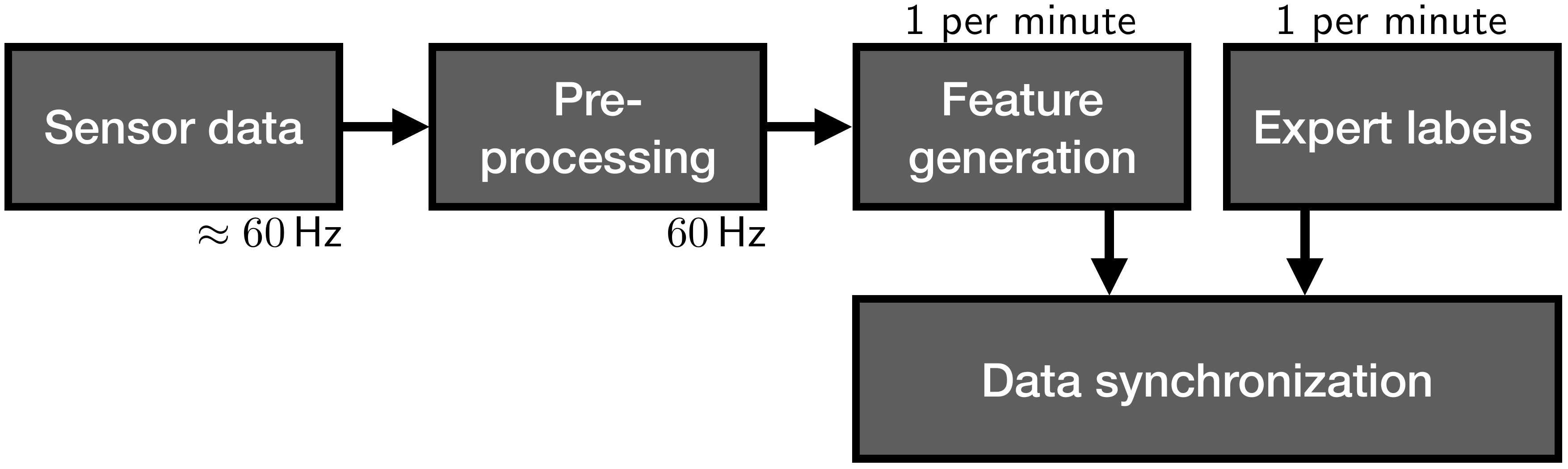}
    \caption{{\bf Inertial data and expert ratings processing.} The sensor data is pre-processed before PD features are extracted. The resulting data are then synchronized with the PD and activity annotations performed by the expert rater.}
    \label{feat_gen}
\end{figure}

The collected inertial data are processed to quantify relevant motion features which are then synchronized with PD and activity annotations, as schematically visualized in Fig.~\ref{feat_gen}. The resulting data set is used for modeling the PD annotations in the hierarchical approach. 

As the raw inertial data includes sensor noise, we smooth the accelerometer and gyroscope data with a two-directional Butterworth band-pass filter
\begin{equation}
    \begin{split}
        \left( \tilde{a}_x, \tilde{a}_y, \tilde{a}_z\right)^\top &= f_\text{Bw} \left( a_x, a_y,a_z, lb, ub \right),\\
        \left( \tilde{\omega}_x, \tilde{\omega}_y,\tilde{\omega}_z\right)^\top &= f_\text{Bw} \left( \omega_x, \omega_y,\omega_z, lb, ub \right).
    \end{split}
\end{equation}
The lower bound~$lb$ of the cut-off frequency was set to 0.1~Hz to filter out sensor drift, and the upper bound~$ub$ was set to 20~Hz to filter out high frequency noise. Furthermore, to avoid dependency of the signal on the wrist band placement (left versus right wrist) and orientation (lateral versus distal and upright versus inverted), the vector norm~$\|\cdot\|$ of both filtered inertial units is calculated,
\begin{equation}
    \begin{split}
        \delta_\text{acc} &= \|\left( \tilde{a}_x, \tilde{a}_y, \tilde{a}_z\right)^\top\| \\
        \delta_\text{gyr} &= \|\left( \tilde{\omega}_x, \tilde{\omega}_y,\tilde{\omega}_z\right)^\top\|,
    \end{split}
    \label{sigfilt}
\end{equation}
and thus scalar signals~$\delta_\text{acc}$ and~$\delta_\text{gyr}$ are obtained. All subsequent feature generation is performed on the processed signals.

\begin{figure}[!t]
    \centering
    \includegraphics[width=\columnwidth]{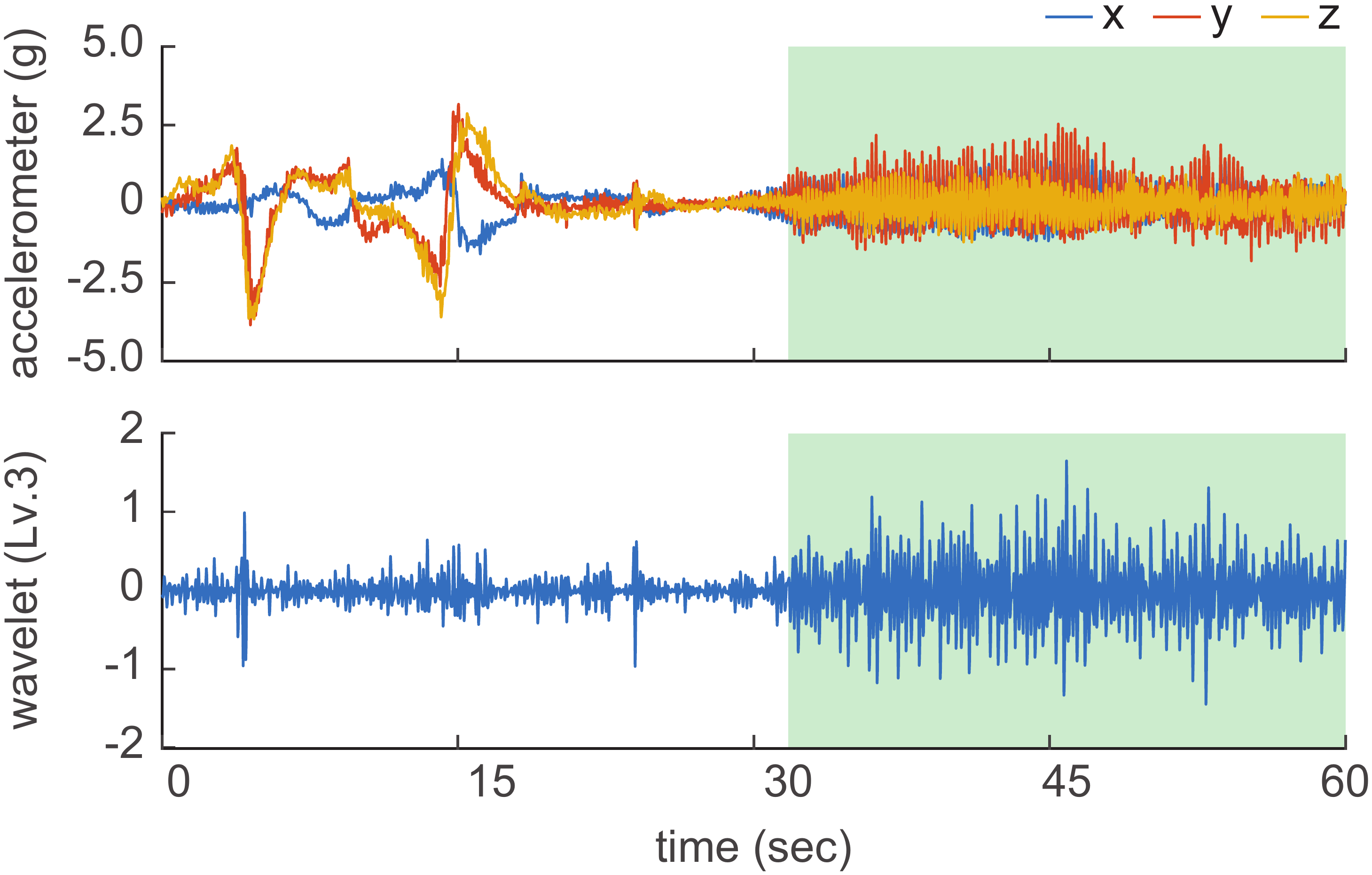}
    \caption{{\bf Wavelet transformation of the accelerometer data.} The discrete wavelet decomposition is performed on the vector norm of 3D accelerometer (and gyroscope) measurements to calculate features that describe characteristics of the 60 seconds time window. Only the third level of the accelerometer signal decomposition is depicted. The green shaded area indicates presence of the tremor according to the PD annotation.}
    \label{feature}
\end{figure}

As demonstrated in Fig.~\ref{activity}, the PD classes have different spectral characteristics. Therefore, we base the feature generation on a time-frequency transformation, namely on wavelet decomposition of the processed sensor data. The signals~$\delta_\text{acc}$ and~$\delta_\text{gyr}$ are transformed using Daubechies wavelets~$\psi_3$ of order 3. The odd-numbered decomposition levels 1, 3, 5, 7 and 9 are employed, as those layers cover the bandwidth of activity levels present in daily living activities according to our observation~(\ref{data}). In Fig.~\ref{feature} a raw accelerometer signal and the third wavelet decomposed level of the corresponding filtered signal vector norm are depicted. The lower part of the figure shows that wavelet decomposition is capable of differentiating voluntary motion (white background) from the tremor-induced motion~(green shaded area). 
We remove the even-numbered layers from the model to minimize redundancy in the feature space.
Then, for each decomposed level~$\tilde{\delta}_{\text{acc},i} = \psi_{3}(\delta_\text{acc},i)$ and~$\tilde{\delta}_{\text{gyr},i} = \psi_{3}(\delta_\text{gyr},i)$, where $i \in \{1,3,5,7,9\}$, characteristic features are calculated. 
The features consist of standard deviation, norm, maximum, root mean square, kurtosis and skewness, as they encode motion properties of the displayed PD class and activity. 
In addition, the signals~$\tilde{\delta}_{\text{acc},i}$ and~$\tilde{\delta}_{\text{gyr},i}$ are differentiated for all $i \in \{1,3,5,7,9\}$ and the standard deviation, norm and root mean square are reapplied to the differentiated signals. A logarithmic scaling is used on a selection of features to improve the activity level separation, as the logarithmizing stretches small positive signals. More specifically, the logarithm is taken of all features obtained from the gyroscope and of the differentiated accelerometer features. 

Every feature is computed for each one minute time window~$t$, if the size of the sample set~$\mathcal{J}_t$ that is captured during window~$t$, contains at least 10\% of the number of data samples that should be captured during one minute, i.e. $\# \mathcal{J}_t \geq 360$ sensor measurements (60~Hz sampling rate times 60~seconds corresponds to 100$\%$). Hence, we only omit data windows that suffer from severe data loss. We allow this small percentage of data samples per captured time window, because the Bluetooth data transmission between the smartwatch and the storage device (smartphone) was occasionally interrupted. 

% still features
The features introduced so far characterize the power in the inertial data and are thus best suitable for describing the tremor and dyskinesia. As bradykinesia is characterized by very slow motion and complete absence of motion, rest phases in the patients' sensor data additionally need to be quantified. 
Therefore, features encoding the length of rest phases within the time window are investigated. We define rest as the proportion of data where the processed inertial signals~(\ref{sigfilt}) are below a given threshold~$c$,
\begin{equation}
    \text{rest}_\text{acc/gyr} = \sum_{j\in\mathcal{J}_t} \frac{\mathbb{B}\left({\delta}_\text{acc/gyr}(j) < c \right)}{ \# \mathcal{J}_t },
\end{equation}
where~$\mathbb{B}(\cdot)$ denotes the boolean operator that assigns the numbers $\{0,1\}$ depending on whether the relation on the inside is false or true. Multiple thresholds are introduced to compensate for inter-patient and inter-activity variability and to cover all severity levels of the PD classes. We use 0.1, 0.15, 0.2, 0.25 and 0.3~$G$ as thresholds $c$ for accelerometer and 1, 1.25, 1.5, 1.75 and 2~$dps$ for gyroscope, respectively. As during severe bradykinesia it can happen that the patient does not move during the whole one minute window, additionally the rest proportion over a 5 minute window~$\bigcup_{\tau \in T}\mathcal{J}_{\tau}$ is calculated, consisting of the two minutes before and after the current time window, $T = \{t-2,t-1,t,t+1,t+2\}$. %As autocorrelation of bradykinesia is high for long time intervals, see~Fig.~\ref{symptom}, the overall condition of the patient is assumed to be the same during those 5 minute windows. 

% PKG
Additionally to the previously mentioned features, we include features that are inspired by the Parkinson's KinetiGraph system~\cite{Griffiths2012}. The two raw inertial signals are filtered, using a stronger bandpass filter (with limits 0.2~Hz and 4~Hz) to keep voluntary motion only. Then, the maximum and the mean spectral power at the maximum are calculated, as those features have been reported to provide promising results for severity estimation of PD classes~\cite{Ossig2016}.

The total number of features obtained is 132 for both inertial sensors (the accelerometer and gyroscope). They are concatenated into the feature vector~$\lambda\in \R^{132}$. We use this vector to model and predict the PD classes.

\subsection{Multi-layer Gaussian Process}
\label{sec:MGP}
In the following, we present our technique to estimate PD class severity $\xi$ from the feature vector $\lambda$.
We propose to use multiple successive GPs to estimate the unknown function
\begin{equation}
    \varphi(\lambda)=\xi.
\end{equation}
GPs are well suited for modeling human movement behavior due to their property to generate smooth motion predictions for nonlinear dynamics. 

The training input set~$\{\x_i\}_{i =1}^\nu$ to the multi-layer GP is the feature vector~$\{\lambda_i\}_{i = 1}^\nu$.
Specifically, at first a \textbf{tremor GP} is trained to recognize the  severity of the tremor from the training input ~$\left\{\lambda_i\right\}_{k = 1}^{\nu}$. The training output set~$\{y_{\text{TM},i}\}_{i = 1}^{\nu}$ is optimized to approximate the severity,
\begin{equation}
    y_{\text{TM},i} \approx 
    \begin{cases}
        \xi_i & \text{if } i\in \mathcal{J}_\text{TM}, \\
        0 & \text{otherwise},
    \end{cases}
\end{equation}
where the set~$\mathcal{J}_\text{TM} := \{j\in \{1,\ldots,\nu\} \wedge \xi_j \text{ contains tremor} \} $.
The next hierarchical level is triggered for the non-tremor data only~$\left\{\lambda_i\right\}_{i = 1, i\notin \mathcal{J}_\text{TM}}^{\nu}$.
This second layer comprises two GP estimations; the \textbf{dyskinesia GP} for modeling dyskinesia, and the \textbf{bradykinesia GP} for modeling bradykinesia. Each of the models is trained to approximate the PD class and severity with its output
\begin{equation}
    \begin{split}
        y_{\text{DK},i} &\approx 
        \begin{cases}
            \xi_i & \text{if } i\in \mathcal{J}_\text{DK}\wedge i\notin \mathcal{J}_\text{TM},\\
            0 & \text{else,}
        \end{cases}\\
        y_{\text{BK},i} &\approx 
        \begin{cases}
            \xi_i & \text{if } i\in \mathcal{J}_\text{BK}\wedge i\notin \mathcal{J}_\text{TM},\\
            0 & \text{else,}
        \end{cases}
    \end{split}
\end{equation}
where the index sets~$\mathcal{J}_\text{DK}$ and~$\mathcal{J}_\text{BK}$ are defined as~$\{j\in \{1,\ldots,\nu\} \wedge \xi_j \text{ dyskinetic}\}$ and~$\{j\in \{1,\ldots,\nu\} \wedge \xi_j \text{ bradykinetic}\}$, respectively.
In the third layer the decision among the balanced, dyskinesia and bradykinesia classes is made based on the results from layer two: When both GP models provide outputs~$y_{\text{DK},i}<\tilde c$ and~$y_{\text{BK},i}<\tilde c$ below a certain threshold $\tilde c$, we consider the correct classification to be balanced. Otherwise the PD class of the GP model providing the higher predicted value is selected.

After the model training is finished in the multi-layer GP, we aim to provide PD class and severity estimates for unseen input data. Given a new feature vector~$\lambda_{\nu+1}$, the process output describes a Gaussian distribution, which has the GP mean prediction as the expected value.
%The individual steps inside the hierarchical algorithm are the following.
%The tremor GP provides in the first layer a severity estimate~$y_\text{TM}$ for the new input $\x_i^\ast$. If no tremor is indicated, predictions~$y_\text{DK}$ and~$y_\text{Bk}$ for the severity of dyskinesia and bradykinesia, respectively, are provided in the second layer, and the decision for one of the PD classes is made in the third layer.
The function~$\lfloor \cdot \rceil$ rounds the GP mean prediction~$\hat{y}_{\nu+1} \in \R$, which is obtained in continuous space, to the nearest integer, and in the unlikely case of predictions outside the scale limits~$[-0.5,4.5)$, maps the negative and positive values to 0 and 4, respectively. It is performed to match the interval variable~$\{0,1,2,3,4\}$ of the PD annotations.
An illustration of the hierarchical approach is provided in Fig.~\ref{hier_model}.

Detailed background information on GP modeling is provided in the Appendix.

\begin{figure}[!t]
    \centering
    \includegraphics[width=\columnwidth]{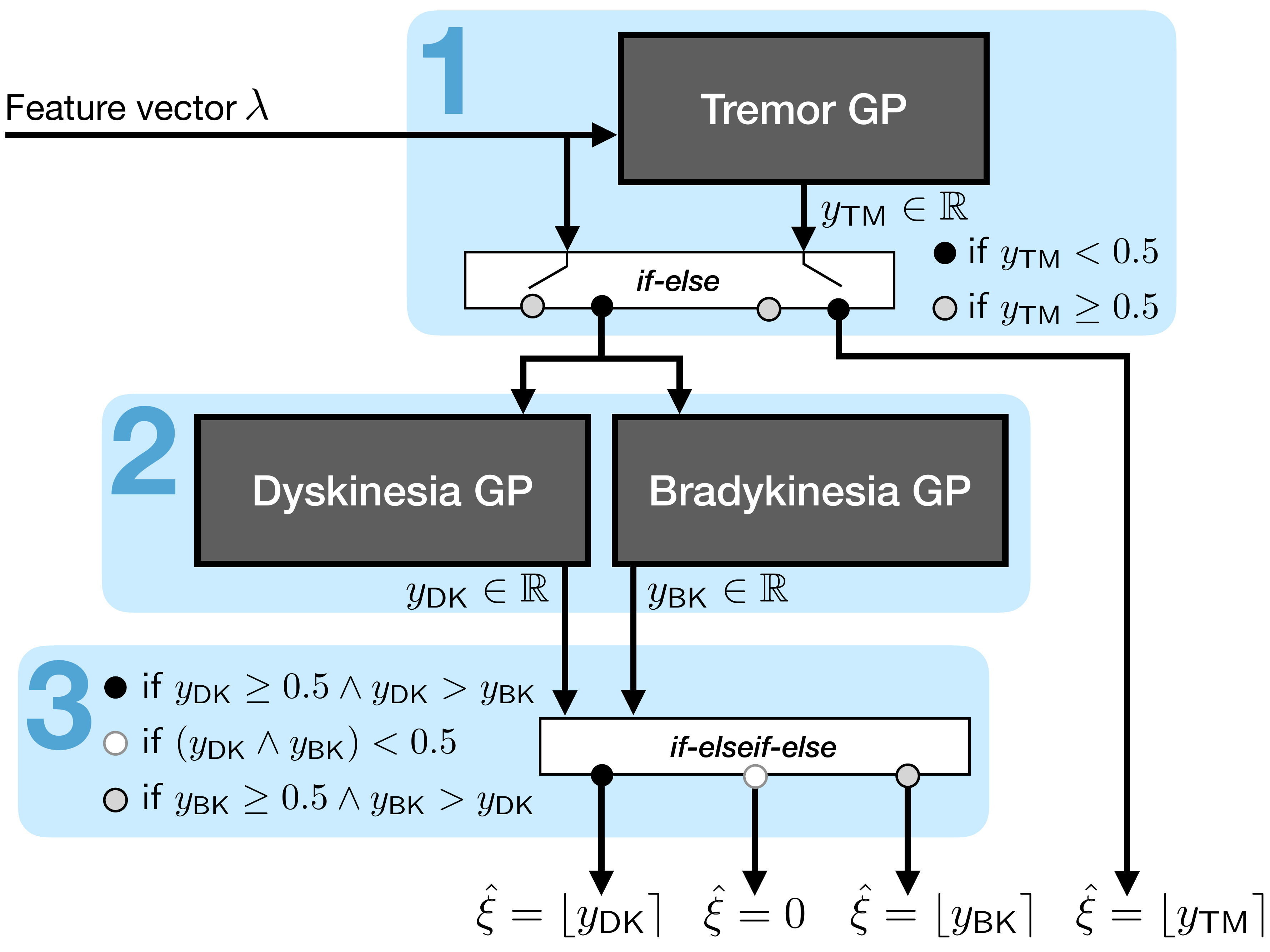}
    \caption{{\bf Hierarchical algorithm scheme. } The feature vector $\lambda$ is the input to the hierarchical approach. 
        After testing for tremor in the first model layer ($y_\textsf{TM} <0.5$ or $y_\textsf{TM} \geq 0.5$), the remaining non-tremor data is tested in the second layer for dyskinesia and bradykinesia severity.
        Each of the models provides an estimate of the severity in the continuous space $\mathbb{R}$, where, however, values outside the interval $[-0.5, 4.49]$ are extremely rare. 
        In the third layer of the hierarchical approach we decide among the three remaining classes bradykinesia, dyskinesia or neither (balanced) depending on the estimated severities across the PD classes. If both predictions obtained from the dyskinesia GP and the bradykinesia GP, respectively, are below a certain threshold ($c = 0.5$), the balanced condition is nominated. If any of the predicted values exceed the threshold, the PD class of the larger value is appointed. To determine one of the interval severity levels 1-4 in the PD annotations, the GP prediction is rounded to an integer value.}
    \label{hier_model}
\end{figure}

%\subsection{Training versus Test Input for GP Models} 
%\input{input/setup}

%\subsubsection*{Tremor estimation}
%\input{input/tremor}
%
%\subsubsection*{Support vector machine}
%\input{input/svm}

%\subsubsection*{Gaussian process}
%\input{input/gp}

\subsection{GP Hyperparameter Characteristics} 
To ensure that the GPs generalize well to unseen data, the GPs should learn the characteristics of the movement features rather than the frequency of the PD class observations in the input data. Therefore, the estimation of false positives (\textit{FP}, i.e. predicting a target when there is none) and false negatives (\textit{FN}, i.e. missing the target) are considered as equally undesired, as this induces impediment of over- and underestimation of a PD class at the same time. 
This means the GP models in Fig.~\ref{hier_model} are required to satisfy the property $FN/FP= 1$, which is achieved by selecting the initial hyperparameters of each of the GP models to approximate this ratio. 
As a GP model that roughly meets the $FN/FP= 1$ property in the initial optimization step is already close to a local optimum and the GP model training employs a gradient descent algorithm, it is unlikely that during training the model deviates from producing estimates where $FN/FP\approx 1$.
Furthermore, a relatively large signal noise~$\vartheta_n$ hinders the overfitting of the GP models to the training data sets and thus, reduces the model's training accuracy, but during testing supports the generalization property.

%%%%%%%%%%%%%%%%%%%%%%%%%%%%%%%%%%%%%%%%%%%%%%%%%%%%%%%%%%%%%%%%%%%%%%%%%%%%%%%%
% Results and Discussion can be combined.
\section{Statistical Analysis}
In this section we detail the practical application of the hierarchical approach to the data set. 
First, we introduce a feature vector reduction, to ensure efficient performance of the multi-layer GP. Then, we present how the whole data set is split into disjoint training and test sets, to analyze the model's ability to generalize to unseen data and unknown patients. Furthermore, we explain how we deal with the non-uniform PD class distribution in the whole data set.
Finally, we provide the initial training hyperparamters and independent training accuracies of the tremor GP,  the dyskinesia GP and the bradykinesia GP.

\subsection{Feature Vector Reduction} 
To reduce the computational complexity of the multi-layer GP, each internal GP model is trained and tested on a subset of the 132 dimensional feature vector only. To determine the informative wavelet decomposition levels a detailed interpretation of the behavioral motion spectrum of the patient data is required:
Decomposition level 1 contains motions with very low frequency ($<$0.2~Hz) which is slower than usual human behavior and thus can be assumed to mainly contain sensor drift. Decomposition levels 3 and 5 cover the frequency range of most intended motions and are thus important to distinguish voluntary motion from PD-specific motor abnormalities. Decomposition level 7 includes frequencies that correspond to fast movements, rarely found in the voluntary motion spectrum of daily living activities of elderly PD patients, but frequently occurring during dyskinetic phases. Decomposition level 9 contains motions of the characteristic tremor frequencies \mbox{(4-6}~Hz).

Therefore, we reduce the feature vector~$\lambda\in \mathbb{R}^{132}$; for the dyskinesia GP we delete the wavelet decomposition levels 1 and 9 for both signals (accelerometer and gyroscope), and for the bradykinesia GP we delete the levels 1 and 7 for both signals. Hence, in each of the GP models the dimensionality of the input vector is reduced by $36 = 2\times 2 \times 9$ dimensions (9 is the number of features calculated in one wavelet decomposition level). 
In the tremor GP the full feature vector~$\lambda\in \mathbb{R}^{132}$ is employed to facilitate separation of the PD class from all other incoming signals. 

\subsection{Disjointed Patient Sets} 
We split the participant cohort, consisting of 30 participants, into two disjoint sets; one group consists of the training dataset and the other group is the test dataset which is used to test the trained model's accuracy. With this approach, we not only quantify the ability of our model for perform regression and prediction, but also demonstrate the model's ability to generalize to datasets of unknown patients. Hence, we introduce an approach that does not require fine tuning on the target patient, but is globally applicable to PD patients. 

Specifically, we perform a leave-one-subject-out (LOSO) approach, where we repeat the training and testing procedure 30 times. In each of the independent runs, the test group consists of one participant and the training is performed on the remaining 29 participants in the cohort. We iterate through the participant cohort so that after the 30 trials the dataset of every participant was once the test set. 

\subsection{Non-uniform Symptom Distribution} 
\begin{figure}[!t]
    \centering
    \includegraphics[width=\columnwidth]{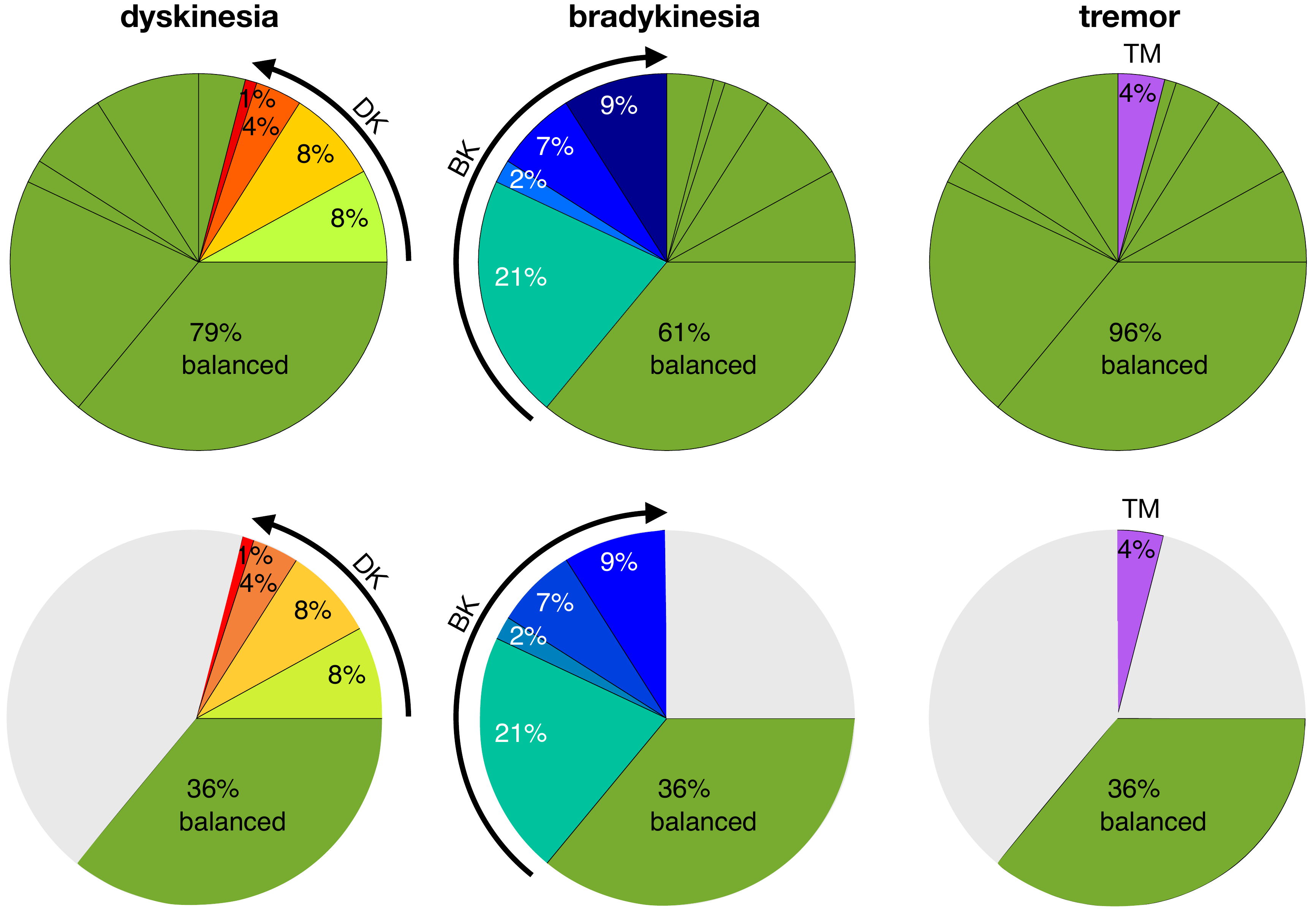}
    \caption{{\bf Distribution of PD class severities according to the different rating scales.} According to the independent scales all data where a specific PD class is not apparent is assigned to the balanced class. To reduce the non-uniformity of the class distribution during the GP model training, the dyskinesia GP, the bradykinesia GP and the tremor GP are trained on the data subsets illustrated in the bottom row left, middle and right, respectively. For all three GPs the model testing, however, is performed on the full data sets (top row).}
    \label{train_set}
\end{figure}

With respect to the three independent rating scales for the PD classes, the whole data set is extremely non-uniformly distributed. In Fig.~\ref{train_set}, top row, the data distributions according to the independent scales are visualized. For instance, according to the tremor rating scale, 96\% of the data does not belong to this class and thus, each of the 30 tremor GPs has an underlying class distribution similar to the pie chart in the upper right corner of Fig.~\ref{train_set}. We say similar, because in each LOSO experiment run a different participant dataset is left out, which slightly effects the class distribution of the training data set of the remaining 29 participants, whereas Fig.~\ref{train_set} shows the PD class distribution of the whole data set. 

Finding suitable GP model hyperparameters that represent the PD class characteristics, not their frequencies, becomes more difficult the more non-uniformly the PD classes are distributed in the training data set. Therefore, we do not train the three GP models on the whole data set, but on data subsets that only comprise the balanced data and the data where the respective class (bradykinesia, dyskinesia or tremor) is present, see Fig.~\ref{train_set}, bottom row. Furthermore, we do not reduce the amount of balanced data, as the balanced class covers the most widespread activities in free living and thus, contains a large diversity of intended motion patterns that need to be distinguished from the unintended motions caused by the movement abnormality. 

The model testing, however, is performed on the whole data set, i.e. the GP predictions of each of the GP models in Fig.~\ref{hier_model} are tested against the PD annotations of the respective test dataset regardless of the PD class. %Recall that only non-tremor data is processed in the second and third layer of the hierarchical approach. 

\subsection{Initial Hyperparameters} 
All 30 training runs per GP model class (tremor, dyskinesia and bradykinesia) are initialized with the same vector of starting hyperparameters~$\theta = (\vartheta_f,\vartheta_l,\vartheta_n)^\top$, in particular for any tremor GP the initial hyperparamter values are~$\theta_\textsf{TM} = (96.83,0.23,0.50)$, for any bradykinesia GP they are~$\theta_\textsf{BK} = (96302550,826659,0.65)$ and for any dyskinesia GP they are~$\theta_\textsf{DK} = (128741,2.26,0.83)$. Those hyperparameter vectors, found by heuristics, fulfill the $FN/FP\approx 1$ property.

\subsection{Training Accuracy of Symptom Severity} 
The three GP model classes in the hierarchical approach are trained independently using the described setup for GP model training. 
We denote with \emph{training accuracy} the percentage of data, where (after the model training has finished) the re-estimated PD severity matches the expert label, and provide the achieved model training accuracies in Table~\ref{training}.
Row-wise, the training accuracy of the tremor GP, dyskinesia GP and bradykinesia GP are provided in terms of the mean and the standard deviation of the 30 independent experiment runs in the LOSO approach. The severity estimation accuracy percentages are given for each of the main activity categories (\textit{sitting, walking, standing, lying} and \textit{other}), and for the total amount of data under \textit{all}. The presented accuracies in all GP models are normalized by the amount of data available in each run, to prevent biased results due to the distortion of repetitions, where the data set for a certain PD class differs strongly in size. During the model training a tradeoff needs to be met between the model adaptation to the training dataset and the model generalizability to new patients. To avoid overfitting of the GP models, the accuracy results of the training are rather low compared to highest achievable training accuracies as we attach importance to the generalizability of the trained models.

\begin{table*}[!t]
    \renewcommand{\arraystretch}{1.3}
    % if using array.sty, it might be a good idea to tweak the value of
    % \extrarowheight as needed to properly center the text within the cells
    \caption{Training Accuracy of the Symptom Severity for each Activity.}
    \label{training}
    \centering
    % Some packages, such as MDW tools, offer better commands for making tables
    % than the plain LaTeX2e tabular which is used here.
    \begin{tabular}{r|c c c c c c}
        Accuracy mean (std) in \% &  other &   sitting & walking & standing & lying & all\\ \hline\hline
        tremor GP &  89.06 (1.24) & 79.75 (1.58) & 82.81 (2.05) & 78.11 (2.83) & 86.03 (1.66) & 83.15 (1.87)\\ 
        bradykinesia GP & 62.56 (3.843) & 60.73 (4.52) & 60.29 (4.26) & 59.80 (3.95) & 60.90 (4.79) & 60.86 (4.28)\\ 
        dyskinesia GP & 49.32 (3.50) & 52.69 (2.16) & 44.55 (1.88) & 46.14 (4.98) & 73.68 (1.24) & 53.28 (1.79)\\ 
    \end{tabular}
\end{table*}

\section{Results}
In this section we report the experimental results of our approach. 
The test accuracies are first independently evaluated for each layer of the multi-layer GP, and then the contingent accuracies over the layers are reported as the total accuracy of the model performance. 
All GP model accuracy results are provided in terms of the mean and standard deviation per main activity category (\textit{sitting, lying, standing, walking} and \textit{other}) to study whether the model prediction was affected by specific activity types.
%Furthermore, we highlight the significance of our dynamical system approach by contrasting the results against pure regression where the GP input for both training and testing is $\x_k= \lambda_k$, i.e. without considering the fluctuation in movement abnormalities as a dynamical system. 
In addition to the \textit{standard accuracy}, the percentage of predictions where the PD annotations 0-4 match the model output, we report the \textit{$\mypm 1$ accuracy}, defining the percentage when the predicted PD severity is at most one level off the severity assessed by the expert rater. Previous studies highlight the agreement rate of PD severity estimation between movement-disorder specialists suffers from the complexity in a display of motor abnormalities and an application of observation-based assessment tools such as UPDRF~(e.g.,~\cite{Heldman2011,Post2005}). As we employed a single rater design to manage the extensive monitoring of PD patients in unscripted activities, the PD severity annotations used in the machine learning expects to have some degree of measurement uncertainty. In the present study, therefore, we also report $\mypm 1$ accuracy as guiding information to show the tendency towards the correct estimation in the face of the inherent input data uncertainty.

The prediction accuracy of the full test data is provided in column \textit{all}, while the accuracies for each of the activity categories \textit{sitting, walking, standing, lying} and \textit{other} are presented in the corresponding column. The results are presented in terms of mean and standard deviation (std) of the accuracies in each of the conducted 30 test runs when iterating through the test patients. The accuracy results are normalized by the amount of data available for each patient and activity to avoid any bias in calculating the mean and standard deviation due to the different data size across the participants. 

%%%%%%%%%%%%%%%%%%%%%%%%%%%%%%%%%%%%%%%%%%%%%%%%%%%%%%%%%%%%%%%%%%
\subsection{Individual GP Model Accuracies}

%In the following, the test accuracies in predicting the severity level of of the three PD classes are provided. 
%
%
%As the symptom distribution among patients is unevenly distributed, the test set includes patients that showed certain symptoms during a single one minute window only. Then the GP prediction can either be right or wrong for that one prediction. However, when calculating the mean and standard over the 30 experiment repetitions, an 0\% or 100\%  more or less has an undesired large effect. 
We start by analyzing the accuracies of each hierarchical layer independently. Hence, the accuracies in the second layer are obtained assuming 100\% correct tremor detection in the first layer. 
In Table~\ref{table_all} the independent in-layer results are presented. 
The severity estimations exceed 75\% accuracy in each activity for the tremor GP and in fewer than 5\% of cases is the predicted severity more than one level off the expert ratings for each activity. Hence, the tremor estimation in layer one is soundly separating the patient data into tremor and non-tremor data, and provides quite accurate predictions of the tremor severity independent of the performed activity.
The captured movements' composition of intended motion and the unintended motor abnormality of dyskinesia and bradykinesia varies from patient to patient, and usually requires individual model tuning to the test patient. Our approach of training the GP models with unified initial hyperparameters, however, is designed to generalize among patients and therefore shows in cases a high standard deviation in the prediction accuracy, indicating a decreased suitability of the GP models for individual patients with atypical movement composition in comparison to the training patient set.
However, the accuracy of the severity estimations in layer two clearly exceeds 50\% on average for bradykinesia and dyskinesia. The $\mypm 1$ accuracy of the severity estimation for bradykinesia and dyskinesia classes exceeds 80\% across all activities. 
The motor abnormalities are exhibited more in some activities than others; the dyskinesia GP performs particularly well in estimating the PD classes and severities during lying, while this is the most difficult activity for the bradykinesia GP. The severity predictions of the dyskinesia GP during lying are more than 98\% correct or by at most 1 level off. 
For the testing results we obtain averaged $FN/FP$ ratios of 1.16, 0.98 and 0.87 for the tremor GP, the bradykinesia GP and the dyskinesia GP, respectively.

% Place tables after the first paragraph in which they are cited.
\begin{table*}[!t]
    \renewcommand{\arraystretch}{1.3}
    \caption{Individual Layer Testing Accuracy of the Symptom Severity for each Activity.}
    \label{table_all}
    \centering
    \begin{tabular}{c l | c c c c c c}
        GP type & accuracy $(\%)$ & other & sitting & walking & standing & lying & all\\ \hline\hline
        \multirow{2}{*}{tremor} 
        & mean (std) 
        & 88.77 (15.72) & 79.13 (24.37) & 81.56 (20.58) & 78.50 (26.53) & 84.97 (15.64) & 82.58 (20.57)\\ 
        & $\mypm 1$ mean ($\mypm 1$ std) 
        & 98.92 (5.17) & 95.53 (9.22) & 97.77 (7.57) & 95.74 (11.93) & 97.52 (6.55) & 97.09 (8.09)\\ 
        % BK
        \rowcolor{gray!20} & mean (std) 
        %           & 52.05 (12.85) & 53.82 (14.35) & 55.70 (14.53) & 48.28 (17.22) & 33.90 (11.61) & 48.75 (14.11) \\ 
        & 70.44 (22.17) & 71.89 (22.29) & 67.24 (25.15) & 49.75 (33.15) & 47.54 (30.81) & 61.37 (23.16)\\
        \rowcolor{gray!20}\multirow{-2}{*}{bradykinesia} & $\mypm 1$ mean ($\mypm 1$ std) 
        %            & 89.10 (6.91) & 90.17 (10.72) & 91.83 (8.16) & 89.22 (10.77) & 77.11 (18.50) & 87.49 (11.01)\\ 
        & 89.28 (16.40) & 89.54 (20.67) & 91.25 (20.18) & 93.63 (12.56) & 82.69 (26.03) & 89.28 (18.18)\\
        % DK
        & mean (std) 
        & 46.99 (18.91)& 53.52 (20.36)& 50.11 (22.47) & 52.21 (24.77) & 85.84 (12.89) & 57.15 (19.87) \\ 
        \multirow{-2}{*}{dyskinesia} & $\mypm 1$ mean ($\mypm 1$ std) 
        & 88.49 (20.63) & 88.61 (16.43) & 89.53 (14.23) & 87.50 (16.09) & 98.31 (3.22) & 90.23 (13.33) \\ 
    \end{tabular}
\end{table*}

%NEW RESULTS WITH HYP: phi_DYS = [28741;2.26] (regress1)

\subsection{Total GP Model Accuracies}
Next, we investigate the \textit{total accuracy} of the GP models, i.e. the probability of both layers (the first and the second) being predicted correctly at the same time. Specifically, the total accuracies are the percentage of accurate predictions of the bradykinesia GP and the dyskinesia GP, respectively, intersected with the accuracy of the tremor GP. 
For simplicity of presentation we provide the accuracy and the $\mypm 1$ accuracy in terms of the percentage of correct predictions of the hierarchical approach for all patients taken together. In total, in 1318 instances of GP predictions tremor is estimated. The remaining 8619 data samples are processed in the second layer of the hierarchical approach. Inside the second layer, the bradykinesia GP and the dyskinesia GP never falsely predicted the presence of both PD classes in parallel for data of the same one minute time window. Hence, the discrimination worked precisely with the proposed motion features.
Table~\ref{class_accuracy} notes the total test accuracy for the tremor, bradykinesia and dyskinesia and for the balanced condition, where movement disorders are absent. The accuracies of the severity estimations (0-4) are provided for the activity categories sitting, walking, standing, lying, other and all. 
For the full data set the tremor severity is estimated precisely in more than 80\%, the bradykinesia severity in more than 60\% and dyskinesia severity in almost 50\% and the balanced condition is detected correctly in more than 36\%. For all PD classes the $\mypm 1$ severity estimation accuracy exceeds 80\% on the whole data, demonstrating a reliable detection of movement abnormalities in PD patients.
\begin{table*}[!ht]
    \renewcommand{\arraystretch}{1.3}
    \centering
    \caption{Total Testing Accuracy of the Symptom Severity for each Activity.}
    \begin{tabular}{c l |c c c c c c}
        PD class & severity $(\%)$ &    other &    sitting &      walking &     standing & lying &       all\\ \hline\hline
        \multirow{2}{*}{tremor} 
        & accuracy & 88.77& 79.13& 81.56&78.50& 84.97& 82.58\\     
        &  $\mypm 1$ accuracy &  98.92& 95.53& 97.77& 95.74& 97.52& 97.09\\   
        \rowcolor{gray!20}
        & accuracy & 66.09 & 67.44 & 58.64 & 43.38 & 45.05 & 61.15\\     
        \rowcolor{gray!20}\multirow{-2}{*}{bradykinesia}  
        & $\mypm 1$ accuracy & 83.52 & 83.54 & 79.50 & 85.54 & 77.33 & 81.99\\     
        \multirow{2}{*}{dyskinesia} 
        & accuracy & 40.80 & 47.41 & 37.13 & 43.38 & 77.48 & 49.24\\     
        & $\mypm 1$ accuracy & 81.95 & 81.67 & 76.27 & 78.68 & 89.80 & 81.98\\    
        \rowcolor{gray!20}
        & accuracy & 34.46 & 39.91 & 19.76 & 23.98 & 48.28 & 36.70\\     
        \rowcolor{gray!20}\multirow{-2}{*}{balanced}  
        & $\mypm 1$ accuracy &  84.74 & 81.24 & 67.38 & 78.32 & 90.85 & 81.06\\  
    \end{tabular}
    \label{class_accuracy}
\end{table*}

\section{Discussion}
The results from the previous section show that our hierarchical approach is capable of estimating PD classes with a good accuracy. We maintain quite precise estimation of the class severity across the presence of various activities including \textit{sitting, walking, standing, lying} and \textit{other}, indicating that our approach is robust against task-oriented arbitrary motions that are non-specific to PD.

In more detail, our model found the activities labeled as standing and lying most difficult for estimating the severity of bradykinesia. This might be due to the absence of large motions in these cases and apparent similarities in behavior between the activity and the motor symptom, including the eyes of the observers who denotes PD annotations. Furthermore, lying frequently is reported for the participants with severe bradykinesia. %During an episode of bradykinesia, standing motion can partially be distorted due to the patients holding onto something.
The severity predictions of the dyskinesia GP during lying, in contrast, are extremely precise. It seems that the much reduced presence of voluntary motions in these activities results in a higher signal-to-noise ratio during dyskinesia estimation.
Detecting the balanced class correctly was particularly difficult during walking, since this activity has an extraordinarily high PSD compared with the other activities~(see Fig.~\ref{psd}) and thus, is likely to be mistaken as tremor or dyskinesia. Note that due to the difference in the training and testing datasets the testing accuracies partly exceed the corresponding training accuracies. The GP models learned the characteristics of the various PD classes and thus, detected data with high accuracy that belongs to a different class than the individual GP model.

The $\mypm 1$ accuracies, which we consider more significant than the accuracies themselves, demonstrate particularly reliable tendency estimations; they exceed 80\% for all activities in the individual layer testing and often are above 90\%. Likewise, in the total accuracy testing the $\mypm 1$ accuracies mostly are above 80\% and only once fall below 75\% (for predicting the balanced class during walking). These results support the importance of predicting PD manifestation with a hierarchical approach using advanced machine learning techniques. 
In comparison to the state of the art in autonomous PD class recognition for PD, we cover a wider range of PD states than most related work by considering tremor, bradykinesia and dyskinesia. Further, our approach allows for state estimation during unscripted daily living activities only with one inertial sensor, and without imposing specific tasks or motion patterns. %Hence, our method significantly decrease the inconvenience to the patient during symptom detection and severity estimation. 

A weakness of the approach concerns the data separation in the first hierarchical layer. Currently, a single GP model is trained for all tremor cases, thus this class does not differentiate whether the patients are co-exhibiting dyskinetic or bradykinetic symptoms. As the tremor is reported as a major initial symptom of PD~\cite{Hoehn1967}, finer classification of the tremor movements will make an important improvement upon the current results. The major hurdle in achieving this with the present study was an observation of a very few cases of tremor with dyskinetic tendency in our patient group. With a sufficient amount of data, estimating of tremor model with a finer class resolution might become possible.

%Regarding the window size of one minute, which was set through the labels provided by the expert rater, a trade-off was met between rating accuracy and feasibility of manual labeling. In future investigation, however, we suggest to either allow variable length windows, that end on symptom state, symptom severity or activity change, or to provide finer time intervals for rapidly changing conditions such as in tremor phases. Determining the optimal window size can be done by randomly cropping the windows to shorter fixed size time segments and finding highest correlations among data segments corresponding to same severity levels. Due to the restriction to solely rating the predominant symptom, multiple subsegments from a single data window are required to counteract the increased noise level induced by cropped segments that might be poorly described by the labeled symptom.

In the current approach we assume false positive (\textit{FP}) and false negative (\textit{FN}) predictions to be equally undesired. In other medical applications, however, other prediction characteristics could be prioritized, which require the adaptation of the initial GP hyperparameters. To give an example of another \textit{FN/FP}-ratio: if detecting all non-balanced PD classes is three times more important than predicting \textit{FP}s, this induces a GP model prediction ratio of $FN/FP \approx 0.33$. An alternative prediction characteristic is for instance: a PD class should be missed in at most 5\% of the estimations. In the present approach, appropriate hyperparameters for such prediction characteristics are found by heuristics. 

\section{Conclusion}
In this article, we introduced a hierarchical approach for autonomous severity estimation of PD states from inertial sensor data, namely for non-tremulous bradykinesia, non-tremulous dyskinesia, and tremor. 
%The essential contribution is the development of a motor symptom severity estimation approach. 
%Hence, the proposed approach can differentiate the patient's symptoms with high precision during daily living, without restrictions on the patient's activities.
We motivate the hierarchical structure of the GP over these classes of PD symptoms by analyzing their class characteristics and propose expressive inertial data features, including those in the time-frequency domain. %, namely wavelet decomposition, enables our approach to achieve accurate estimation results during unstructured patient activities. 
%Moreover, as we rely on a commercially available low-cost wearable sensor, handling of the tracking device is affordable, user-friendly and maintenance is minimal.
%In the experimental evaluation, we used a PD patient data set to analyze the estimation accuracy of non-tremulous bradykinesia, non-tremulous dyskinesia, and tremor. Importantly, we estimated their severity levels with respect to the standard PD scales. 
The results showed reliable estimation of symptom severities, and suggest that the proposed approach can differentiate the patient's symptoms with good precision during daily living, without imposing specific activities on patients. %; the categories in which the activities were subdivided comprise \textit{walking, sitting, standing, lying} and \textit{other}. 

\appendix[Gaussian process]
A Gaussian process (GP) model is a data-driven approach that convinces owing to its robustness in describing dynamics from mere observations. It is a nonparametric regression method defined in continuous space. We employ GP regression for its suitability to model non-linear mappings and for its natural ability to model predictive conditional probabilities including a best estimator and a prediction confidence. 
%This uncertainty estimate depending on the similarity of the test data and the training samples is highly valuable in safety-critical applications. 

Compactly stated, \textit{a Gaussian process is a collection of random variables, any finite number of which have a joint Gaussian distribution}~\cite{Rasmussen2006GP}. Additionally, we give a more explicit formulation.
\begin{defn}
    Let~$\mathcal X$ be a (multidimensional) index set, and denote by~$\{\varphi({\x}) \}_{\x \in \mathcal X}$ a real-valued stochastic process over~$\mathcal X$. Such a process is called Gaussian, if and only if any finite collection of random variables~$\{\varphi({\x_1}),\, \ldots,\, \varphi({\x_\nu})\}$, where~$\nu \in \N$, is $\nu$-dimensional multivariate Gaussian.
    \label{GPdef}
\end{defn}

To provide an intuitive understanding of the process, we explain the process functioning in the following. First, a GP model is trained to approximate an unknown scalar-valued mapping~$\varphi$. The learning comprises the optimization of process hyperparameters, to best describe the mapping from input~$\{\x_i\}_{i =1}^\nu$ to output~$\{y_i\}_{i =1}^\nu$. Hence, our training data set~$\left\{ (\x_i,y_i) \right\}_{i = 1}^\nu$ consists of input-output pairs, where the input can be multidimensional but the output~$y_i = \varphi({\x_i})$ has to be scalar. Then, the GP model returns for any new input value~$\x^\ast \in \mathcal X$ an estimate~$y^\ast$ for~$\varphi(\x^\ast)$ in form of a predictive Gaussian distribution $\No (\mu_{\x^\ast},\sigma_{\x^\ast})$.
An important GP characteristic is shown by~\cite{Rasmussen2006GP}, which states, a GP is fully specified by a mean function $m(\x)$ and a kernel function $k(\x,\x^\prime)$
\begin{equation}
    \begin{split}
        m(\x) &= \E[\varphi(\x)], \\
        k(\x,\x^\prime) &= \E[(\varphi(\x)-m(\x))(\varphi(\x^\prime)-m(\x^\prime))].
    \end{split}
\end{equation}
Thus, we can write
\begin{equation}
    \varphi(\x) \sim \text{GP}(m(\x),k(\x,\x^\prime)),
    \label{GP}
\end{equation}
meaning that the unknown underlying function~$\varphi(\x)$ has the same distribution as our learned GP model and hence, can be approximated by it.

In many applications the mean function~$m(\x)$ is set to zero,~$m(\x) \equiv 0_{\varphi(\mathcal X)}$ as this reduces the computational complexity without limiting the expressive power of the process~\cite[Chap.~6.4.1]{Bishop2006}. Hence, the heart piece of GP modeling concerns the selection and optimization of the kernel function. Apart from few exceptions, the kernel functions can be divided in two classes: stationary and dot-product kernels. Dot-product kernels are based on the magnitude of the samples, while stationary kernels are based on the distance between samples and consequently, are invariant to translation in input space, which is an often desired property. The most prominent example of this second kernel class is the squared exponential kernel function,
\begin{equation}
    k(\x,\x^\prime) = \vartheta_f^2 \exp\left( \frac{\| \x-\x^\prime\|}{2\vartheta_l^2} \right)+\vartheta_n\delta(\x,\x^\prime),
\end{equation}
where~$\delta$ describes the Kronecker delta function and~$\theta = (\vartheta_f,\vartheta_l,\vartheta_n)^\top > 0$ the hyperparameter vector.

\section*{Acknowledgment}
The work of S.H., L.M. and S.E. was supported by the EU seventh framework programme FP7/2007-2013 within the ERC Starting Grant Control based on Human Models (con-humo), grant agreement no. 337654. The work of T.T.U. and D.K. was supported in part by Canada's Natural Sciences and Engineering Research Council and Ontario's Early Researcher Award. The position of U.F was supported by an unrestricted research grant from the Deutsche Stiftung Neurologie and the Deutsche Parkinson Vereinigung.

\ifCLASSOPTIONcaptionsoff
  \newpage
\fi




\balance
\bibliographystyle{IEEEtran}

\bibliography{IEEEabrv,all_refs}

\end{document}